\pdfoutput=1

\documentclass[11pt]{article}

\usepackage[preprint]{acl}

\usepackage{times}
\usepackage{latexsym}

\usepackage[T1]{fontenc}

\usepackage[utf8]{inputenc}

\usepackage{microtype}

\usepackage{inconsolata}

\usepackage{graphicx}

\usepackage{tikz}

\usepackage[utf8]{inputenc} 
\usepackage[T1]{fontenc}    
\usepackage{hyperref}       
\usepackage{url}            
\usepackage{booktabs}       
\usepackage{amsfonts}       
\usepackage{nicefrac}       
\usepackage{microtype}      
\usepackage{xcolor}         
\usepackage{tcolorbox}
\usepackage{amsmath}
\usepackage{amsthm}
\usepackage{graphicx}
\usepackage{mathtools}
\usepackage{cleveref}
\usepackage{algorithm}
\usepackage[noend]{algpseudocode} 
\usepackage{wrapfig}

\usepackage{algorithm}
\usepackage{algorithmicx}
\usepackage{algpseudocode}
\DeclarePairedDelimiterX{\infdivx}[2]{(}{)}{#1\;\delimsize\|\;#2}

\theoremstyle{definition}

\usepackage{mathrsfs}
\usepackage{amsmath}
\usepackage{bm}
\usepackage{multirow}
\usepackage{adjustbox}
\usepackage{graphicx}
\usepackage{colortbl}
\usepackage{arydshln}
\usepackage{microtype} 
\usepackage{amsmath, amssymb, mathtools}
\usepackage{appendix}
\usepackage{wrapfig}
\usepackage{xcolor}
\usepackage{tabularray}
\UseTblrLibrary{siunitx}
\definecolor{Gray}{gray}{0.9}
\definecolor{LightCyan}{rgb}{0.88,1,1}
\usepackage{multirow}
\definecolor{mygray}{gray}{0.6}
\definecolor{cvprblue}{rgb}{0.21,0.49,0.74}

\newcommand{\gray}[1]{\textcolor{gray}{#1}}

\title{Diversifying the Expert Knowledge for Task-Agnostic Pruning \\ in Sparse Mixture-of-Experts}

\author{\thanks{Work done during the internship at Microsoft Research.}Zeliang Zhang$^{1}$\, Xiaodong Liu$^{2}$\, \ Hao Cheng$^{2}$\,\ Chenliang Xu$^{1}$\, Jianfeng Gao$^{2}$\\
$^{1}$University of Rochester\quad
$^{2}$Microsoft Research\quad
\\\footnotesize{\texttt{\{zeliang.zhang, chenliang.xu\}@rochester.edu,}} \footnotesize{\texttt{\{xiaodl, cheng.hao, jfgao\}@microsoft.com}}}

\begin{document}

\maketitle
\renewcommand{\thefootnote}{\fnsymbol{footnote}} 

\begin{abstract}
\label{sec:abst}
In this work, we address the memory overhead of deploying Mixture-of-Experts (MoE) architectures in Large Language Models (LLMs). While MoE layers improve LLM performance without increasing inference costs, the ever-growing number of experts inflates memory requirements, hindering practical deployment. 
Our empirical study reveals that some experts encode redundant knowledge during pre-training.  We thus propose a method of grouping and pruning similar experts to improve the model's parameter efficiency. We validate the effectiveness of our method by pruning three state-of-the-art MoE architectures, including Mixtral, Deepseek-MoE, and Qwen. The evaluation shows that our method outperforms other model pruning methods on a range of natural language tasks. 
\end{abstract}
\section{Introduction}
\label{sec:intro}

Large Language Models (LLMs) have achieved outstanding performance across various tasks by learning a large number of model parameters on large amounts of data, as shown by the scaling laws~\citep{kaplan2020scaling}.  In addition to increasing the depth of neural network models, widening neural networks by using the sparsely-activated mixture-of-experts (MoE) architecture is also proved effective. MoE widens the feed-forward network (FFN) layer (one expert) by having multiple parallel FFNs (experts). During forward propagation, only a subset of these experts is activated. Thus, compared to dense models, MoE models achieve better end-task performance and generalize better to new tasks without increasing computation costs. Notable examples of MoE models include Switch Transformer~\citep{fedus2022switch}, Mixtral-MoE~\citep{jiang2024mixtral}, and Uni-MoE~\citep{li2024uni}. 

Despite significant progress in developing wider and deeper MoE LLMs, the increased memory consumption due to larger model sizes (i.e., increased number of experts) poses a substantial challenge to the deployment of these models in real-world settings. For example, storing and loading Mixtral-8x7B, which has 8 experts in each of its 32 layers, requires approximately $88$ GB. The MoE layers constitute the majority of the parameters. Adding or removing even one expert in each layer can significantly impact overall memory cost and model performance. For example, ~\citep{lu2024not} shows that randomly dropping $2$ experts in each MoE layer reduces the memory cost by $21$ GB, and decreases model performance by $7\%$ on the MMLU benchmark~\citep{hendrycksmeasuring}. 
In this study, we strive to seek the best trade-off between memory efficiency and task performance by identifying an optimal set of experts in each MoE layer to prune.

\begin{figure}
    \centering
    \includegraphics[width=\linewidth]{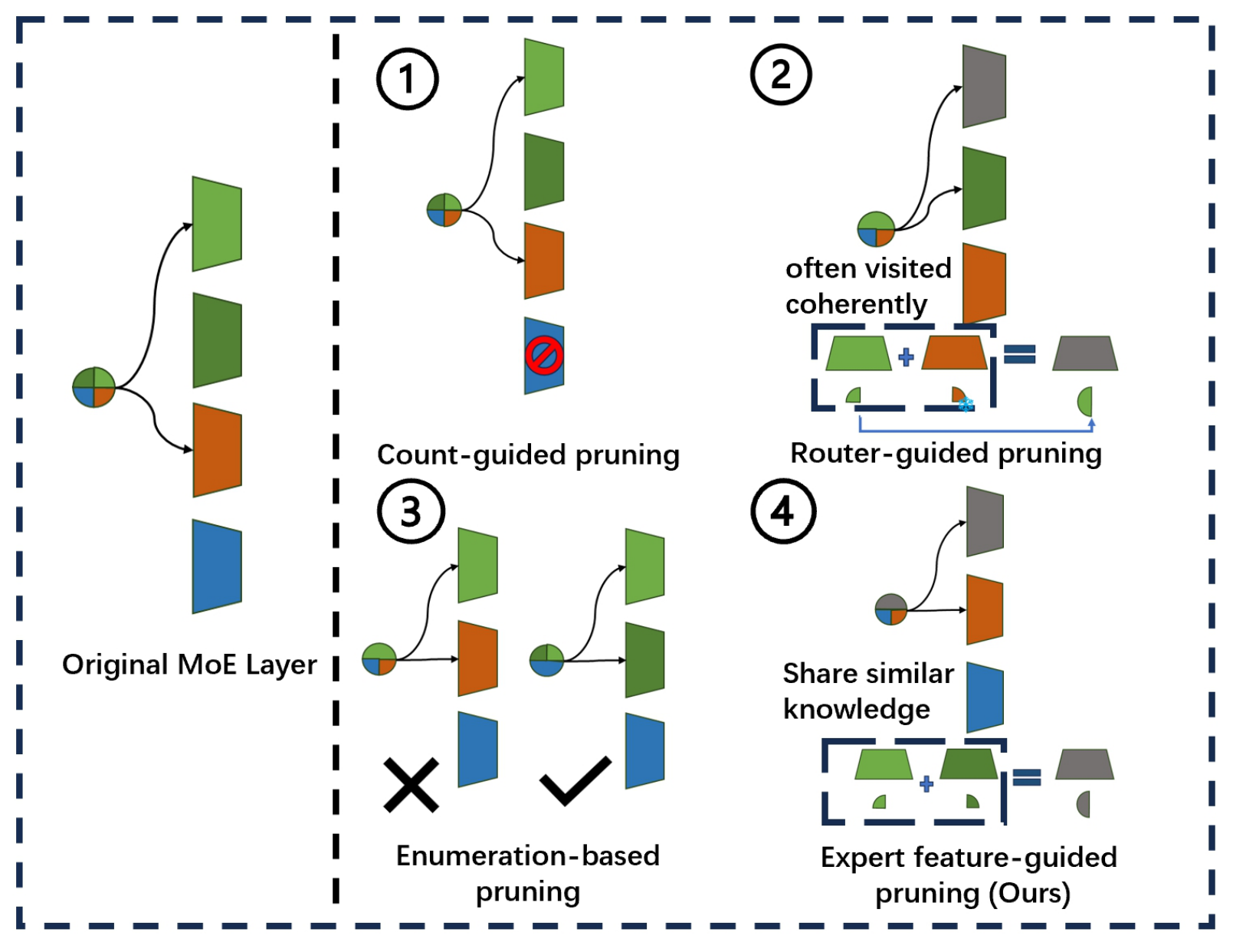}
    \caption{Removing several experts from the original MoE layer would not cause model collapse but improve efficiency. Experts with similar colors share the similar knowledge with each other.  Prior works often utilize expert access information to filter out unimportant experts. In our work, we first group different experts with similar knowledge in the feature space, then merge them along with the routers to prune the MoE layer. This post-processing approach allows us to diversify the features of each MoE layer, thereby preserving the knowledge of the original large models as much as possible while reducing computation and storage consumption.}
    \label{fig:teaser}
    \vspace{-0.5cm}
\end{figure}

There have been several studies on pruning MoE models. 
One line of work utilizes task-specific information to prune irrelevant experts. For example, \citet{chen2022task} prune the less frequently visited experts based on experiments on a range of tasks. \citet{chowdhury2024provably} find that less important experts usually exhibit smaller changes in routing weights during the fine-tuning stage. \citet{li24merge}  merge experts that are frequently visited by tokens of a fine-tuned dataset for pruning.
{While effective, these methods depend on knowing the target tasks. In contrast, task-agnostic pruning methods that do not rely on task information are more appealing and useful in real-world applications because they can apply to both seen and unseen tasks. However, this is more challenging since there are no explicit task and data cues to guide which experts are redundant. }
\citet{he2024demystifying} explore pruning experts with less visited frequency in a task-agnostic calibration dataset but report a significant performance drop. In comparison, \citet{lu2024not} enumerate all the combinations of experts and prune some to achieve a minimum loss of reconstruction, which significantly improves performance. We illustrate the difference among these works in \cref{fig:teaser}. Although pruning MoE models in task-agnostic settings is of great practical value, this area has not been fully explored.

In this work, we explore how to prune MoE models in a task-agnostic fashion. 
Our study is motivated by the finding that, given the same input, many experts respond similarly, indicating that these experts likely encode similar knowledge, and thus are somewhat redundant. 
We propose a method to improve model parameter efficiency by pruning redundant experts in two stages. As shown in \cref{fig:method}, we first identify and group similar experts in the feature space. 
Then, for each group, we merge experts in the weight space to diversify the knowledge in different MoE layers. 
We validate the effectiveness of our method by pruning experts for three state-of-the-art MoE architectures, including Mixtral (Mixtral-8x7B and Mixtral-8x22B), Deepseek-MoE (Deepseek-MoE-16B) and Qwen (Qwen2-57B-14A).
The evaluation shows that our method outperforms other model pruning methods on a range of natural language tasks. Our contributions are summarized as follows,

\begin{enumerate} 
\item We empirically validate that some experts within each well-trained MoE layer encode similar knowledge, making them somewhat redundant. 
\item We propose a two-stage, task-agnostic method for grouping and merging redundant experts, which is further divided into data-centric and model-centric implementation strategies. 
\item We demonstrate the effectiveness of our method by pruning experts from a series of state-of-the-art MoE models, including Mixtral-MoE, DeepSeek-MoE, and Qwen-MoE. The results from a greedy search for MoE pruning further validate the success of our approach. \end{enumerate}


\section{Related Work}
\noindent \textbf{Sparse MoEs}. The Mixture-of-Experts (MoE) structure is firstly applied in classical machine learning models by \citet{jacobs1991mixtures} and \citet{jordan1994hierarchical}, then widely used in various deep learning models~\citep{yuksel2012twenty,masoudnia2014mixture,zhang2023robust}. Recently, some works employ the MoE to scale the capacity of transformer-based models, especially the large language models~\citep{shazeer2017outrageously,lepikhin2020gshard,zoph2022st}. It adapts the original large feed-forward network (FFN) in each transformer block into multiple smaller FFNs, forming an expert layer with a router that computes the weighted output of each MoE layer. Sparse MoEs were first proposed by \citet{fedus2022switch}. In this approach, only a few experts are activated in each layer, accelerating training and inference while significantly increasing the number of parameters for greater model capacity. Many sparse MoEs have been developed and open-sourced within the AI community, such as Switch Transformer~\citep{fedus2022switch}, Mixtral-8B~\citep{jiang2024mixtral}, and Uni-MoE~\citep{li2024uni}. Recent studies also indicate that neural networks with the MoE structure exhibit better generalization ability compared to dense models~\citep{zhu2022uni,li2022sparse}.

\noindent \textbf{Model Pruning}. Model pruning involves removing unimportant parameters from a well-trained neural network to balance task performance and computational efficiency~\citep{liu18rethinking, wang2020pruning,liu2021discrimination}. Pruning techniques can be categorized into unstructured pruning~\citep{liao2023can,shi2024towards,mason2024makes}, which introduces sparsity in the weight matrix by setting some parameters to zero, and structured pruning~\citep{lemaire2019structured,fang2023depgraph,shen2022prune}, which removes entire neurons, layers, or blocks, reducing redundancy and being more suitable for acceleration on GPUs~\citep{choi2021gpu}. Many efforts have been made to leverage model pruning techniques to reduce the memory consumption of neural networks, spanning a range of models from conventional architectures like CNNs~\citep{luo2018thinet}, RNNs~\citep{zhu2017prune}, and LSTMs~\citep{ding2020prune} to modern large models such as Llama~\citep{xia2023sheared} and Stable-Diffusion~\citep{castells2024ld}.

While the large amount of parameters in sparse MoEs benefits the model's capacity to achieve good performance at the pre-training stage, the increasing memory consumption causes great challenges to fine-tuning different downstream tasks. In this paper, we work on pruning the sparse MoEs to reduce redundant experts in the task-agnostic setting, which enhances the computational and memory efficiency throughout the fine-tuning process, and scalability of deploying these models.

\section{Method}
\label{sec:meth}
\vspace{-0.2cm}
In this section, we present our approach for pruning experts within MoE layers in large language models. Our goal is to identify experts that share highly similar knowledge and merge them, thereby reducing the model size and improving efficiency without significantly degrading performance.

\subsection{Notation and Preliminaries}
\label{subsec:not}
We denote $F(\cdot;\Theta,W, K)$ as an MoE layer.
Here, $\Theta=\{\theta_1, \theta_2,...,\theta_N\}$ is the set of parameters for $N$ experts $\{f_{n}(\cdot;\theta_{n})\}_{n=1}^{N}$, where each expert is typically a feed-forward network (FFN). The routing matrix $W\in\mathbb{R}^{N \times d}$ determines which experts are selected for each token. K denotes how many top experts are chosen for each input token.

For a token $x\in \mathbb{R}^{d}$, we first compute the routing logits to measure how well it matches each expert: $p_n(x) = \frac{e^{W_{n}x}}{\sum_{t=1}^{N}e^{W_{t}x}}$. Then, we select top-$K$ experts, $\{f_{n}(\cdot;\theta_{n})\}_{n=i_1}^{i_K}$, based on these logits. Finally, we compute the MoE layer's output as the weighted combination of the chosen experts: $y = \sum\limits_{n=i_1}^{i_K}p_n(x)\cdot f_{n}(x;\theta_{n}).$


\subsection{Task Definition}
\label{subsec:task}
Previous works~\citep{lu2024not,he2024demystifying,li24merge} have shown that pruning some experts in MoE layers can improve model efficiency in both inference and fine-tuning without causing model collapse. However, an open question remains: which experts should be removed?

Formally, consider a LLM $\mathcal{M}(\cdot;\mathcal{F})$ comprsing $L$ MoE layers. Each MoE Layer $F^{l}(\cdot; \Theta^l, W^l, K)$ has $N$ experts.
Our objetive is to only remain $r$ experts from each MoE layer:
\begin{equation}
\small
\begin{aligned}
    \hat{\Theta}^l = \Theta^l \setminus \{\theta^l_{s^l_1}, \theta^l_{s^l_2}, \ldots, \theta^l_{s^l_{N-r}}\} \\
    \hat{W}^l = W^l \setminus \{W^l_{s^l_1}, W^l_{s^l_2}, \ldots, W^l_{s^l_{N-r}}\}
\end{aligned}\label{eq:rmR}
\end{equation}

\noindent We aim to find these indices $s^l = \{s^l_1, s^l_2, \ldots, s^l_{N-r}\}$ that minimizes loss on a generic dataset $D$:
\begin{equation}
\small
    \begin{aligned}
        \min\limits_{(x,y)\sim\mathcal{D}}\mathcal{L}(\hat{\mathcal{M}}(x;\hat{\mathcal{F}}),y).
    \end{aligned}\label{task:find_experts}
\end{equation}
The search space \((C^{r}_{N})^L\) is enormous, making direct combinatorial search intractable.

\subsection{Measuring Expert Similarity with CKA}\label{sec:motivation}
\label{subsec:cka}

\begin{figure*}[htbp]
    \centering
    \includegraphics[width=\linewidth]{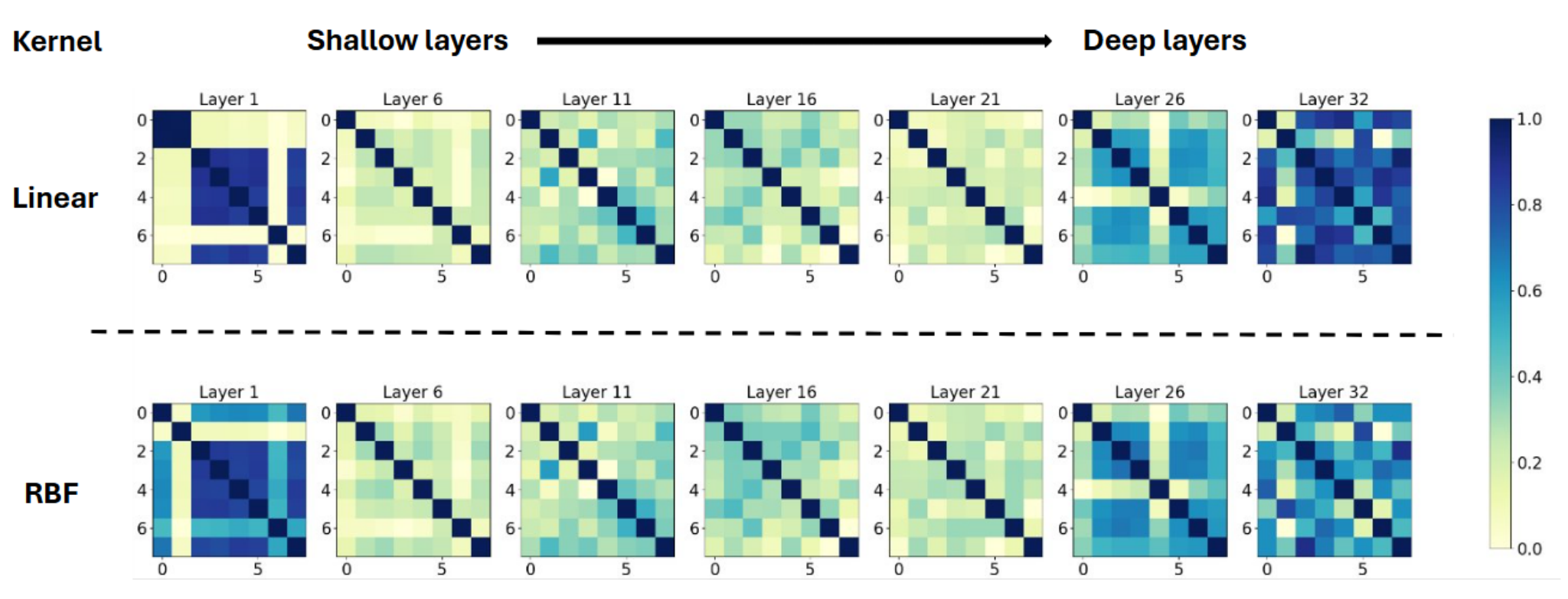}
    \caption{Evaluation of the expert similarity for different MoE layers in Mixtral-8x7B under two kernel-based CKA criteria (Linear and RBF). A darker color indicates a greater similarity between experts. }
    \label{fig:motivation_mixtral_experts}
    \vspace{-0.5cm}
\end{figure*}

Our key insight is that experts exhibiting similar behaviors likely contain redundant knowledge. By identifying and pruning these redundant experts, we can simplify model and improve its efficiency. 

To quantify similarity, we use Centered Kernel Alignment (CKA)~\citep{kornblith2019similarity,davarireliability,smerkousenhancing} as the criteria to evaluate the similarity between experts in each MoE layer. Intuitively, CKA measures how similarly two experts transform a shared batch of inputs. For any two experts  $f_i$ and $f_j$, given a batch of inputs $x=\{x_1, x_2, ...,x_{s}\}$, the similarity $\rho_{ij}$ is computed as follows, \vspace{-0.2cm}
\begin{equation}
\small
    \begin{aligned}
       \rho_{ij} = \text{CKA}(K^i,K^j) = \frac{\text{HSIC}(K^i,K^j)}{\text{HSIC}(K^i,K^i)\cdot\text{HSIC}(K^j,K^j)},
    \end{aligned}
    \label{eq:cka}
\end{equation}
where $K^i$ and $K^j$ are kernel matrices constructed from experts' outputs on the same input batch $x$, and HSIC is the Hilbert-Schmidt Independence Criterion~\citep{greenfeld2020robust}. The definition is as follows: $\text{HSIC}(K^i,K^j)=\frac{1}{(s-1)^2}\text{tr}(K^iHK^jH)$, $K^{i}_{mn}=k(f_i(x_m),f_i(x_n))$, $H=I - \frac{1}{s}11^T$, and $k(\cdot,\cdot)$ is the kernel function. Notably, all experts are provided with the same input batch—no token distribution by the router function is involved—allowing for a clearer representation of the distinct knowledge each expert acquires within the same layer.

We evaluate the similarity between experts in Mixtral-8Bx7B using 32 randomly selected samples from the C4 pre-training dataset~\citep{raffel2020exploring}. Following \cref{eq:cka}, we compute expert similarity with both a linear kernel (equivalent to using a dot product) and an RBF kernel.

As shown in \cref{fig:motivation_mixtral_experts}, darker cells indicate greater similarity between pairs of experts. Both kernel measures reveal similar patterns across layers, indicating that some experts are moderately to highly similar. This suggests the feasibility of pruning certain MoE layers. For example, in the first layer, experts 2 through 5 consistently show similarity scores above 0.7, suggesting that removing one of these experts would likely have minimal impact on overall model performance. Full evaluation results can be found in \cref{sec:apped_results}.

\subsection{Discovering and Merging Similar Experts}
\label{subsec:alg}
\begin{figure*}[htbp]
    \centering
    \includegraphics[width=0.8\linewidth]{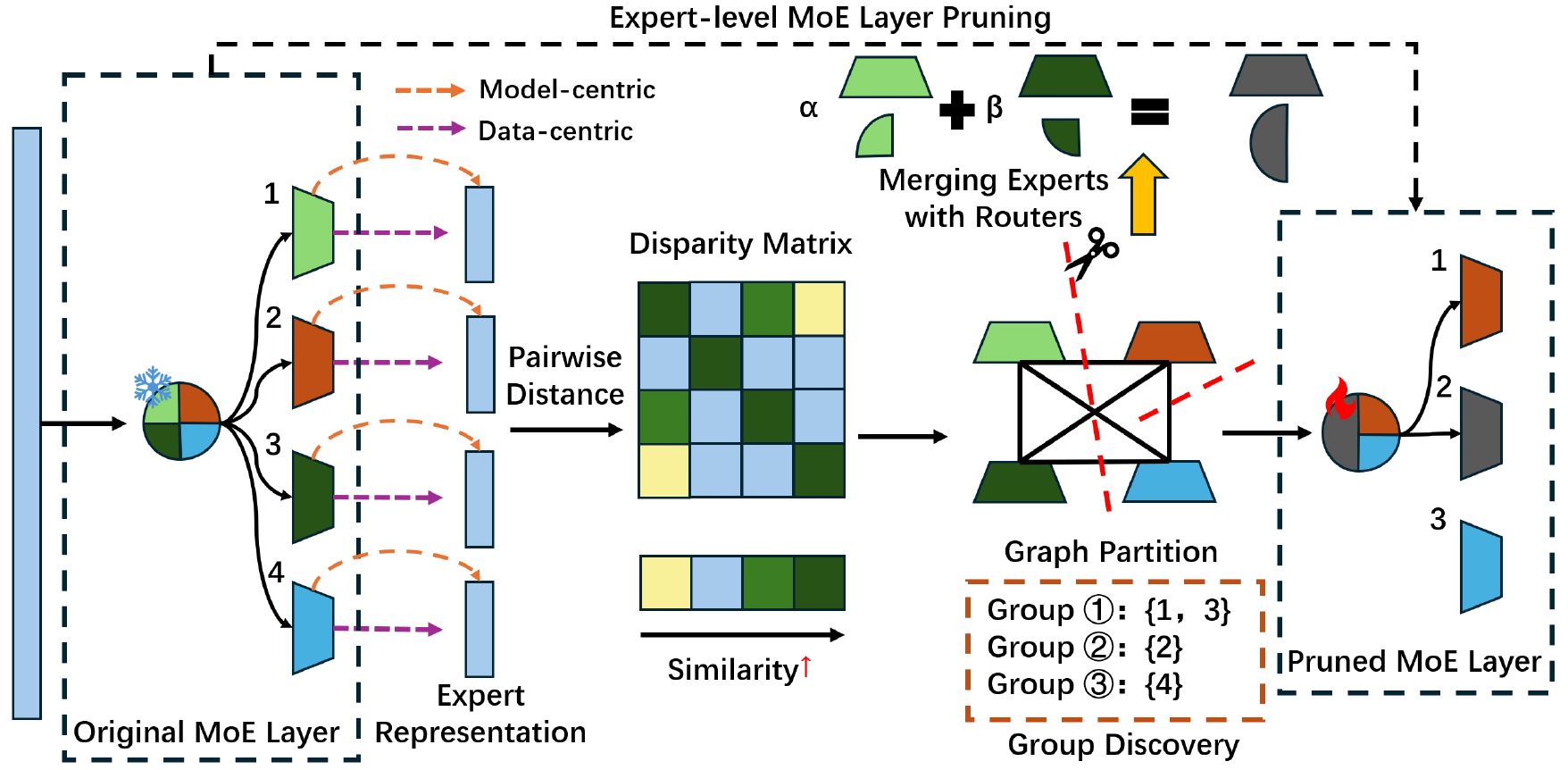}
    \caption{We first leverage model or data-centric strategies to obtain the expert representation, then compute the pairwise distance to get the disparity matrix. Based on the expert similarity matrix, we can group similar experts with shared knowledge in the same cluster, which can be merged on the weight space for pruning. }
    \label{fig:method}
    \vspace{-0.5cm}
\end{figure*}

As shown in \cref{fig:method}, our pruning strategy consists of two key steps to reduce the number of experts from $N$ to $r$: (1) identifying groups of similar experts and (2) merging each group into a single expert. 


\noindent \textbf{Discovering Similar Experts from the Expert Graph}.
We construct a unidirectional graph $\mathcal{G}=\{\mathcal{V}, \mathcal{E}, \mathcal{A}\}$ to cluster similar experts, where $\mathcal{V}$ is the set of nodes (each representing an expert), $\mathcal{E}$ denotes the set of edges connecting experts that exhibit positive similarity, and $\mathcal{A}$ is a weight matrix that encodes these pairwise similarities. The procedures for constructing $\mathcal{A}$ is shown as follows. 


First, we represent each expert $f_{i}$ by its output embedding on a shared calibration dataset (data-centric) or by its own weights (model-centric) if data is unavailable, as $\mathcal{R}(f_{i})$. 


Second, we calculate the weight on each edge of $\mathcal{E}$ to represent the similarity between experts, \textit{i.e.}, $\mathcal{A}\left(\mathcal{E}\right)$, using CKA (data-centric) as \cref{eq:cka}  or mean squared error (model-centric)\footnote{More discussion is provided in Appendix \ref{append_code}}. Each element $\mathcal{A}\left(\mathcal{E}_{ij}\right)=\text{CKA}(\mathcal{R}(f_{i}),\mathcal{R}(f_{j}))$ reflects how similar experts $f_i$ and $f_j$ are in the transformed space $\mathcal{R}(f_{})$ as the previous step.

Last, we split $\mathcal{G}$ into $r$ subgraphs $\{\mathcal{G}_{i}\}_{i=1}^{r}$ to group similar experts. The experts indexed by $\mathcal{V}_{i}$ of $\mathcal{G}_{i}$ share the most similar knowledge with each other and show much difference with experts indexed by $\mathcal{V}_{t}$ of $\mathcal{G}_{t}$, $t \neq i$. This can be formulated as the follows,  
\begin{equation}
\small
    \begin{aligned}
        &\max \sum_{i=1}^{r} \left(\sum_{j,k \in \mathcal{V}_{i}} \mathcal{A}\left(\mathcal{E}_{jk}\right) - \sum_{t \neq i}^{r} \sum_{j \in \mathcal{V}_{i}, k \in \mathcal{V}_{t}} \mathcal{A}\left(\mathcal{E}_{jk}\right)\right), \\
        &\textit{s.t.} \quad \bigcup_{i=1}^{r} \mathcal{V}_{i} = \mathcal{V}, \quad \mathcal{V}_{i} \cap \mathcal{V}_{t} = \emptyset \quad \text{for} \quad i \neq t.
    \end{aligned}\label{eq:find}
\end{equation}
 Our objective is to minimize the intra-group similarity (first term) and maximize the inter-group difference (second term) based on the pairwise similarity $\mathcal{A}\left(\mathcal{E}_{ij}\right)$ based on expert $i$ and $j$.

\noindent \textbf{Merging Similar Experts}
To diversify  the experts and preserve the different knowledge learned by different models clustered in the same group~\citep{wortsman2022model,wang2022adamix,Xinyu2024modelglue}, we merge the clustered  experts with their routers on the weight space as follows,
\begin{equation}
\small
    \begin{aligned}
        \hat{\theta}_{n} &\leftarrow \sum_{i=1}^{|\mathcal{V}_{n}|}\alpha_{i}\theta_{\mathcal{V}_{n}(i)}, ~~ \hat{W}_{n} &\leftarrow \sum_{i=1}^{|\mathcal{V}_{n}|}\alpha_{i}W_{\mathcal{V}_{n}(i)},
    \end{aligned}
    \label{eq:model_soup}
\end{equation}
where $\mathcal{V}_{n}$ is the set of similar experts in the $n$-th cluster $\mathcal{G}_{n}$, and  we have $\sum_{i=1}^{|\mathcal{V}_{n}|}\alpha_i=1$. We update the MoE layer $F$ by respectively replacing all the experts grouped in $\mathcal{G}_{n}$ by a single FFN expert layer $f(\cdot;\hat{\theta}_{n})$ and corresponding routing weights $\hat{W}_{n}$, $n=1,2,...,N-r$. The pseudo-code is provided in \cref{append_code}.

\subsection{Practical Considerations}\label{sec:approach}



\noindent \textbf{Computing the disparity matrix $\mathcal{A}(\mathcal{E})$}. We propose two strategies for computing the disparity matrix $\mathcal{A}(\mathcal{E})$: a data-centric strategy and a model-centric strategy.

For the \textbf{data-centric} strategy, since the full pre-training datasets for different models are large and inaccessible, we use the C4 dataset, a commonly used smaller pre-training subset that serves as an effective surrogate for capturing task-agnostic knowledge of experts~\cite{lu2024not}. Specifically, by disabling the router function, for the same input, we use the output of each expert as its expert representation $\mathcal{R}(\cdot)$. Then, we compute the expert similarity and discover the similar experts by \cref{eq:find}.  To enhance generalization and mitigate overfitting to the selected samples during model pruning, we apply data augmentation by randomly mixing token embeddings during the representation computation at the discovery stage.

For \textbf{model-centric} strategy, we propose two ways to prune models with only expert weights, which encode the dataset information during the training process and can be a good agency for expert representation. One is to leverage the vectorized weights directly ($\clubsuit$).  The other is to leverage the local linearity\footnote{In short, the 'local linearity' property of neural networks refers to the observation that, despite having multiple hidden layers and non-linear activation functions, neural networks exhibit linear behavior within a local region. This property motivates our design of a single-layer approximation using the surrogate weight.}  of neural networks~\citep{zhang2020empirical} to compute the surrogate weight matrix ($\spadesuit$).

Taking the FNN $f(\cdot;\theta)$ in Mixtral-8x7B as an example, it consists of three linear layers, \textit{i.e.}, $\theta=\{\theta_1,\theta_2, \theta_3\}$. For input $x$, the output is computed as $f(x;\theta)=\theta_{2}(\sigma(\theta_{1} x) \cdot \theta_{3} x)$. Then, the vectorized weight ($\clubsuit$) is $\text{concat}\{\theta_1, \theta_2, \theta_3\}$, while the surrogate weights can be obtained by $\theta_2(\theta_1\cdot\theta_3)$ ($\spadesuit$). Compared to vectorized weights, \textit{surrogate weights are more flexible with model size, showing more stable performance} (see \cref{append_sup:more_results}).

\textit{The selection of distance function}. While the CKA metric provides an accurate estimation of expert similarity, it can be both memory- and computation-intensive. As an alternative, cosine similarity may be used as an efficient approximation. This corresponds to computing CKA for a single sample (data-centric), replacing the $K_i$ and $K_j$ in \cref{eq:cka} with the weights of the $i$-th and $j$-th experts, respectively (model-centric). In practical deployments, we also observe that mean squared error performs well across several models, including DeepSeek-MoE and Qwen-MoE.

\textit{Discussion on two strategies}. Ensuring distributional similarity between surrogate and real datasets is essential for the data-centric method to accurately estimate expert similarity; without the guarantee, its performance could be degraded significantly. While pre-trained data for many public MoEs, such as Mixtral, DeepSeek, and Qwen, is unavailable, our study, along with prior research, demonstrates that leveraging a small subset of the open-source dataset C4 can still achieve competitive performance. Alternatively, the model-centric approach presents a robust solution, eliminating the need for data access during pruning and, in some cases, achieving performance comparable to or exceeding that of data-centric methods. More discussion is provided in \cref{append:discuss_two_strategies}.

\noindent \textbf{Merging Strategies and Trade-offs}. We consider three approaches in practice deciding $\alpha$ to merge similar grouped experts.  The first is to only maintain the experts with the maximum visiting frequency and drop all the others. The second is to set $\alpha_{i}=\frac{1}{|\mathcal{V}_{n}|}$, which uniformly assembles all the experts grouped. The last one is learning $\alpha$ to merge the grouped experts by minimizing the following loss function,
 \begin{equation}
 \small
     \begin{aligned}
     &\mathcal{L}(\{\alpha_n\}_{n=1}^{|\mathcal{V}_n|})=\Vert 
y-F(x;\hat{\Theta},\hat{W},K) \Vert,\\
&\textit{s.t.}~\hat{\Theta}(n)=\lambda(\sum_{i=1}^{|\mathcal{V}_{n}|}\alpha_{i}\theta_{\mathcal{V}_{n}(i)}), \hat{W}_{n} = \sum_{i=1}^{|\mathcal{V}_{n}|}\alpha_{i}W_{\mathcal{V}_{n}(i)},
     \end{aligned}
     \label{eq:loss_learn_alpha}
 \end{equation}
where the ground-truth is the output of original MoE layer $y=F(x; \Theta, W, K)$, and we jointly optimize $\lambda$ and $\alpha$ for different merging groups. Among these three strategies, both the first and third require the presence of data, while the second is compatible with both data-centric and model-centric approaches. We empirically find that the uniform souping strategy offers more stable performance and greater efficiency in task-agnostic model pruning (see \cref{append_sup:more_results}
). While the learning strategy can yield slightly better performance, it is more time-consuming due to the need for tuning the parameter $\alpha$.

\section{Evaluation}

\begin{table*}[]
\centering
\caption{Results on pruning the Mixtral-8x7B from 8 experts to 6 and 4 experts in each MoE layer.  The first and second columns respectively indicate the results of the pruned model with $6$ and  $4$ experts.}
\label{tab:main_result_7b_sim}
\resizebox{\textwidth}{!}{
\begin{tabular}{cc|cccccccc}
\toprule
\multicolumn{2}{c|}{Dataset} & \multicolumn{4}{c}{MMLU}                   & \multirow{2}{*}{BoolQ} & \multirow{2}{*}{OpenBookQA} & \multirow{2}{*}{RTE} & \multirow{2}{*}{Average} \\
\multicolumn{2}{c|}{Method}  & humanities & social science & stem & Other &                        &                             &                      &                          \\\hline
\multicolumn{2}{c|}{\gray{Mixtral-8x7B~\citep{jiang2024mixtral}}}  &            \gray{60.5}         & \gray{77.8}               & \gray{58.9}     &    \gray{74.2}   &      \gray{85.4}                  &         \gray{34.4}                   & \gray{71.1}                     &    \gray{66.0}                     \\
\multicolumn{2}{c|}{Router-guided~\citep{li24merge}}  &            51.8/24.8         & 60.5/26.5               & 46.9/24.7     &    60.5/25.0   &      82.6/39.9                  &          32.0/11.6                   & 70.4/50.9                     &             57.8/29.1              \\
\multicolumn{2}{c|}{Count-guided~\citep{he2024demystifying}}      &            49.2/36.9         &      59.7/45.6          &  45.0/35.1    &  58.2/43.4     &        77.2/76.6                 &    \textbf{33.0}/26.4                         &   56.6/55.9 &              54.1/45.7            \\
\multicolumn{2}{c|}{Enumerate~\citep{lu2024not}}      &            52.4/43.5          &   66.4/52.7             &   49.0/40.4   &  63.7/43.5     &          84.0/80.8              &                  32.6/28.8           &  \textbf{71.1}/66.4                    &  59.9/50.8                        \\\hline
\multirow{2}{*}{\textbf{Ours}}    & \textbf{Model-centric} & 54.4/\textbf{48.1}       &    70.2/\textbf{58.5}            &  51.8/\textbf{45.2}    &   66.8/\textbf{55.2}    &                85.6/83.7        &     31.4/26.2        &       68.9/62.4            &  61.3/54.2                        \\
                         &\textbf{Data-centric} & 56.0/48.0                          &     \textbf{73.1}/57.0 &    \textbf{52.4}/43.3   &          \textbf{68.2}/54.6              &    \textbf{86.4}/83.3           &         31.4/28.5      &         \textbf{69.3}/\textbf{67.1}             &     \textbf{62.4}/\textbf{54.5}
   \\\bottomrule
\end{tabular}}
\end{table*}
\subsection{Experiment Setup}
\noindent  \textbf{Studied Models}. We take pruning three MoE-based architectures as an example, including the Mixtral, Deepseek, and Qwen. For the Mixtral architecture, the Mixtral-8x7B has $32$ sparse MoE-involved layers, in each there are $8$ experts.  The Mixtral-8x22B is similar to Mixtral-8x7B but with $56$ sparse MoE layers.   During the inference, each token will select $2$ experts in each MoE layer. The deepseek model has $28$ layers, and there are $64$ experts in each layer. Each token will pass $2$ shared experts and select $6$ experts during the inference.  The Qwen model also has $28$ MoE layers with $64$ experts in each layer but will activate $8$ experts during the inference.   We include more experimental details in \cref{append:exp_details}.


\noindent \textbf{Pruning Methods}. We take three advanced MoE pruning methods~\citep{he2024demystifying, li24merge, lu2024not} as our baseline for comparison. Among them,  \textit{router-guided} merging~\citep{li24merge} is initially designed for task-specific MoE pruning, where we set the target dataset as the samples from the pre-training dataset. For task-agnostic methods, we select the frequency-based pruning method Expert Trimming~\citep{lu2024not}, also the \textit{count-guided} strategy and loss-based pruning method~\citep{lu2024not}, namely the \textit{Enumerate} in our paper. Under the task-agnostic pruning setting, we disable all the fine-tuning stage of all methods for fair comparison.  

For our method, we respectively report the best results of \textit{our data-centric} and \textit{model-centric} methods in pruning models. Following the previous setting~\cite{lu2024not}, we use $128$ samples in C4 for computation in data-centric pruning methods, while the model-centric method doesn't rely on the data.  More detailed results on different implementations of discovery and merging steps (\textit{e.g.}, vectorized and surrogate weight strategy at the discovery step, max or learning strategy at the merging step) are provided in \cref{append_sup:more_results}.

\begin{table*}[]
\centering
\caption{Results on pruning the  Mixtral-8x22B from 8 experts to 6 and 4 experts in each MoE layer. The first and second columns respectively indicate the results of the pruned model with $6$ and  $4$ experts.}
\label{tab:main_result_22b_sim}
\resizebox{\textwidth}{!}{
\begin{tabular}{cc|cccccccc}
\toprule
\multicolumn{2}{c|}{Dataset} & \multicolumn{4}{c}{MMLU}                   & \multirow{2}{*}{BoolQ} & \multirow{2}{*}{OpenBookQA} & \multirow{2}{*}{RTE} & \multirow{2}{*}{Average} \\
\multicolumn{2}{c|}{Method}  & humanities & social science & stem & Other &                        &                             &                      &                          \\\hline
\multicolumn{2}{c|}{\gray{Mixtral-8x22B~\citep{jiang2024mixtral}}}  &            \gray{68.6}         & \gray{84.1}               & \gray{67.1}     &    \gray{78.7}   &      \gray{87.9}                  &         \gray{0.358}                   & \gray{71.2}                     &    \gray{70.4}                   \\
\multicolumn{2}{c|}{Router-guided~\citep{li24merge}} &  27.3/22.7   & 25.4/25.8 & 24.4/24.0 & 27.9/23.4 & 62.8/62.7  & 12.8/13.0 & 54.2/49.5 & 33.5/31.6           \\
\multicolumn{2}{c|}{Count-guided~\citep{he2024demystifying}} & 58.0/45.7 & 74.9/57.7 & 54.1/42.0 & 70.2/45.7 & 81.5/74.4 & 35.2/27.0 & 69.3/57.4 &  63.3/50.0           \\
\multicolumn{2}{c|}{Enumerate~\citep{lu2024not}} & 60.4/53.9 & 78.0/67.2 & 59.5/52.3 & 73.0/64.2 & 87.4/80.5 &   35.0/31.1 &  70.1/67.9  & 66.2/59.6           \\\hline
\multirow{2}{*}{\textbf{Ours}}                     
   & \textbf{Model-centric} &\textbf{63.7}/\textbf{58.1} & \textbf{80.0}/\textbf{72.5} & \textbf{62.1}/\textbf{54.3} & \textbf{75.6}/\textbf{68.3} & 88.0/\textbf{85.2} & 34.6/31.2 & 69.0/\textbf{68.6} & \textbf{67.6}/\textbf{62.6} \\
   & \textbf{Data-centric}  & 62.3/57.8 & 78.5/69.7 & 60.2/51.3 & 73.4/64.2 & 87.6/83.1 & \textbf{35.8}/\textbf{33.2} & \textbf{71.1}/68.1   & 67.0/61.1               
   \\ \bottomrule
\end{tabular}}
\end{table*}

\noindent \textbf{Evaluation Datasets}. The open-sourced Language Model Evaluation Harness library~\citep{gao2021framework} is used to evaluate the performance. We select four  tasks, including MMLU~\citep{hendrycksmeasuring}, BoolQ~\citep{clark2019boolq}, OpenBookQA~\citep{mihaylov2018can}, and RTE~\citep{bentivogli2009fifth}. Among these tasks, MMLU is the most challenging one, which consists of $57$ subtasks, where we present four groups, namely the humanities, social science, stem and other. 



\subsection{Pruning the Mixtral Architecture}
We present the results of the pruning of the Mixtral architecture, including Mixtral-8x7B and Mixtral-8x22B. Both have $8$ experts in each MoE layer. We apply different methods to prune them from 8 experts to 6 and 4 experts in each layer, respectively. The results of the two models are shown in \cref{tab:main_result_7b_sim} and \cref{tab:main_result_22b_sim} respectively.

\noindent \textbf{Results on Mixtral-8x7B}. We can see that all our proposed four strategies surpass the related works, with a clear margin performance improvement of $1.5\%$ on average. Compared with count-guided strategy~\cite{he2024demystifying} which just drops the experts less visited, router-guided strategy~\citep{li24merge} has a large improvement of $3.7\%$ on average by merging these experts, showing that \textit{the merging operation plays a crucial role in preserving the expert knowledge}. Besides, compared with count-guided and enumerate strategies~\citep{lu2024not} which all adopt the dropping strategy,  we can see that  \textit{directly leveraging the expert feedback rather than the routing frequency is more suitable for task-agnostic pruning in MoE layers}.   We can also notice that the model-centric method surpasses all the other data-involved pruning baseline methods. This suggests that \textit{weights already encode fruitful data information and can be deployed to group experts for pruning}.



\noindent \textbf{Results on Mixtral-8x22B}. In this experiment, the model-centric approach achieves the best result, with only a minor performance drop of $2.8\%$ on average compared to the full model. Our proposed data-centric method ranks second to last, with a performance gap of $0.8\%$ compared to the runner-up method. This suggests that model-centric approaches exhibit better robustness when pruning models of different scales, while data-centric methods are more prone to overfitting on small calibration datasets (as evidenced by the collapse of the route-guided method in \cref{tab:main_result_22b_sim}).



\subsection{Pruning the Deepseek Model}
\begin{figure}
    \centering
    \includegraphics[width=\linewidth]{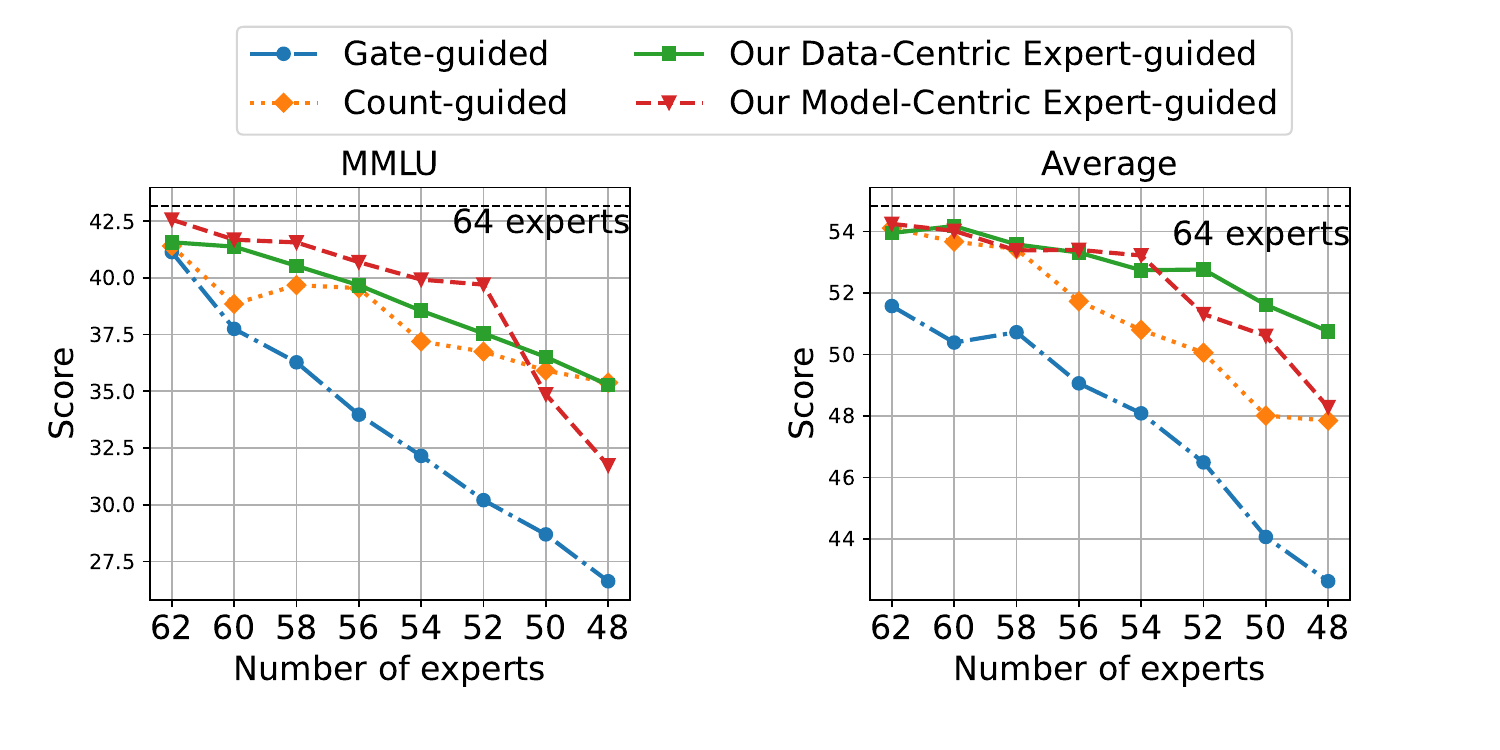}
    \caption{Results on pruning the Deepseek-MoE-16B with different strategies.}
    \label{fig:deepseek}\vspace{-0.5cm}
\end{figure}

We also evaluate the effectiveness of our proposed pruning method on compressing the DeepSeek architecture. Unlike Mixtral-MoE, DeepSeek-MoE features a shared expert and incorporates more fine-grained experts at each MoE layer. Specifically, we pruned the \textit{non-shared experts} of DeepSeek-MoE-16B, reducing the number of experts from 64 to 48 using various model pruning strategies. The results are illustrated in \cref{fig:deepseek}\footnote{We  only show the pruning results of the Deepseek and Qwen models on the most challenging MMLU task and the average performance across MMLU, BoolQ, OpenBookQA, and RTE. Full results can be found in supplementary.}. 

Notably, even after pruning one-third of the experts in Deepseek-MoE-16B, our data-centric strategy maintains an impressive average performance of $50.9\%$, with only a $3.1\%$ reduction in performance compared to the full model. In the evaluation of the most challenging MMLU task, our model-centric strategy demonstrates superior performance in most cases, particularly when reducing the number of experts from $62$ to $52$. It consistently outperforms the runner-up baseline method, achieving a clear performance advantage of $2.6\%$.

\subsection{Pruning the Qwen Model}
\begin{figure}[htbp]
    \centering
    \includegraphics[width=\linewidth]{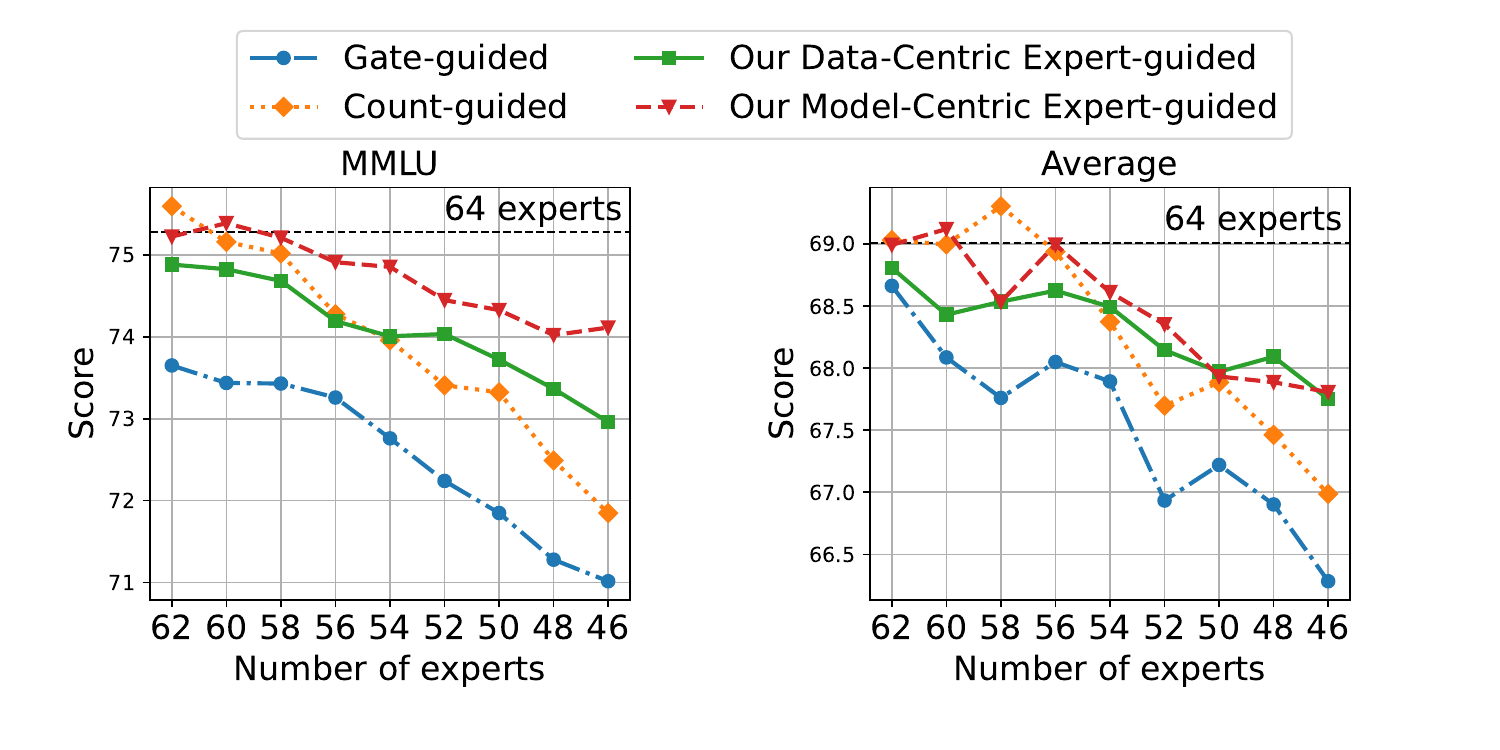}
    \caption{Results on pruning the Qwen2-57B-14A with different strategies.}
    \label{fig:qwen}
\end{figure}
The Qwen architecture is also utilized in our experiments. We study the Qwen2-57B-14A to evaluate the performance of different model pruning strategies. We prune the experts from $64$ experts to $48$ experts in each layer and evaluate the performance on different tasks. 

Reported in \cref{fig:qwen}, when pruning $\frac{1}{3}$ experts of the full model,  the data-centric strategy delivers the best performance, outperforming both our model-centric method and the count-guided baseline method. On the MMLU task, our model-centric strategy achieves approximately $1.5\%$ better performance compared to the runner-up one.

\paragraph{Performance discrepancy between Deepseek and Qwen.} The difference in performance appears to stem from variations in model architectures and training strategies. Specifically, Deepseek-MoE-16B uses shared experts to learn a broad range of knowledge, which can lead to highly diverse expert functions. Consequently, identifying similar experts for pruning becomes challenging, and removing them risks losing unique knowledge—resulting in notable performance drops. In contrast, Qwen2-57B-14A initializes its expert weights from a pre-trained dense model, increasing expert similarity post-training. Meanwhile, Deepseek-MoE-16B’s experts are randomly initialized and trained from scratch, resulting in fewer interchangeable experts. Pruning such independently specialized experts leads to a more pronounced performance decline.

\begin{table}[]
    \centering
    \caption{Evaluation results when ranging the number of samples for expert similarity estimation. The augmentation is disabled in this experiment.}
    \label{tab:diff_samples}
    \resizebox{\columnwidth}{!}{
    \begin{tabular}{c|ccccc}
    \toprule
        \# of samples & MMLU & BoolQ & OpenBookQA & RTE & Avg.  \\\hline
        128 & 61.5  & 85.6  & 32.8  & 68.6   & 62.1 \\
        256 & 61.4 & 85.4  & 33.2  & 69.7   & 62.4\\
        512 & 61.7  & 85.4  & 33.4  & 70.4   & 62.7 \\\hline
    \end{tabular}}
    \vspace{-0.2cm}
\end{table}


\subsection{Ablation Study and Discussion}

\noindent \textbf{Comparison with the greedy search}. To further demonstrate the superiority of our proposed expert pruning strategy, we employ a greedy search approach to identify the optimal candidates in each MoE layer for pruning and compare the results with those obtained using our method. We enumerate all possible combinations of removing two experts in each layer, evaluate the pruned models on the MMLU task, and record the best combination for each layer. We then merge these results to prune the model across all layers. The model pruned using the greedy search approach achieves a performance of $62.22\%$, while our data-centric strategy achieves $61.19\%$, showing a comparable result. Full results are shown in \cref{append_sup:enum}.

\noindent \textbf{On the used samples}. 
We conduct experiments to study the effect of both sample size used for the calibration dataset and augmentation in data-centric pruning methods. As shown in \cref{tab:diff_samples}, increasing the number of samples from $128$ to $512$ \textit{without augmentation} leads to a noticeable performance improvement of up to $0.6\%$.In contrast, when comparing this result with \cref{tab:main_result_7b_sim}, where $128$ samples were used with augmentation, we observe an average performance improvement of $0.3\%$. This demonstrates the importance of both sample size and augmentation in enhancing generalization during the pruning process, as discussed in \cref{sec:approach}.



\noindent \textbf{On merging the routing policy}.The motivation of our work is to diversify expert knowledge by merging similar experts. A key distinction of our approach is the merging of routing policies, which potentially directs more tokens to the resulting merged expert. To highlight the importance of merging routers, we evaluate the performance of our pruning method without merging the routing policies. As shown in \cref{tab:not_merge_router}, compared with the results in \cref{tab:main_result_7b_sim} and \cref{tab:main_result_22b_sim},this results in performance degradation across all tasks and different numbers of experts, underscoring the crucial role of simultaneously merging both routers and experts.

\begin{table}[]
    \centering
    \caption{Evaluation results of our data-centric pruning method without merging the routers.}
    \label{tab:not_merge_router}
    \resizebox{\columnwidth}{!}{
    \begin{tabular}{c|ccccc}
    \toprule
        \# of experts & MMLU & BoolQ & OpenBookQA & RTE & Avg.  \\\hline
        4 & 49.1  & 83.3  & 27.4  & 66.4   & 56.5 \\
        6 & 60.4 & 84.7  & 31.2  & 67.5   & 60.9\\
        \hline
    \end{tabular}}
\end{table}





\noindent \textbf{Efficiency improvement after MoE pruning}\label{append_effi}
To demonstrate the benefits of MoE pruning, we provide a statistical efficiency analysis for Mixtral-8x7B and DeepSeek-MoE-16B. We set the batch size to 8 and evaluate the performance on the MMLU task using the lm\_eval library~\citep{gao2021framework}. Other settings are consistent with the previous experiment.  As shown in \cref{tab:effi_ana}, pruning two experts from each MoE layer in the Mixtral model achieves a 1.17× speedup and a 16.8\% reduction in GPU memory usage. Pruning 16 experts from each MoE layer in the Deepseek model reduces inference time by 15.5\% and GPU memory usage by 14.1\%.  These results  highlight the computational benefits of expert pruning across different model scales.  {More analysis can be found in \cref{append_hint}.} 
\begin{table}[]
\centering
\caption{Computation cost of evaluating Mixtral-8x7B and DeepSeek-MoE-16B on the MMLU task. Experiment settings are consistent with our paper.}
    \resizebox{0.7\columnwidth}{!}{
    \begin{tabular}{c|c|c|c}
    \toprule
      Model &\# Experts & Time (s) & Mem (GB) \\\hline
      \multirow{3}{*}{Mixtral-8x7B}  & 8 (Ori.)	& 281	& 125  \\
       & 6           &  241  & 104  \\
       & 4           &  223  &  83  \\\hdashline
       \multirow{5}{*}{DeepSeek-MoE-16B} & 64 (Ori.) &	457&	64  \\
       & 60	&446	&61  \\
       & 56	&429	&59  \\
       & 52	&413	&57 \\
       & 48	&386	&55 \\\bottomrule
    \end{tabular}}
    \label{tab:effi_ana}
\end{table}

\section{Conclusion}
\label{sec:conc}
In this paper, we work on the task-agnostic pruning of sparse MoEs. We propose discovering similar experts at the feature level and then merging them in the weight space for MoE pruning while preserving as much original expert knowledge as possible. This approach allows the MoE layer to maintain diverse experts with different knowledge, thereby efficiently reducing redundancy.



\section{Limitations}
Several unexplored questions remain in our project. First, we designed various strategies to prune the MoE, and we observed that different models require different strategies to achieve optimal post-pruning performance. It remains unclear what causes these performance differences across strategies. Second, while the learning strategy at the merging step can bring slightly performance improvement, the cost is also large. The question of how to efficiently find the optimal merging coefficients remains. Third, in our work, we prune the same experts across different MoE layers, despite each layer having varying levels of redundancy. A key question remains: how can we push MoE compression to its limits while maintaining acceptable performance?

\bibliography{custom}


\newpage
\appendix
\renewcommand\thefigure{\Alph{section}\arabic{figure}} 
\renewcommand\thetable{\Alph{section}\arabic{table}}    
\setcounter{figure}{0}  
\setcounter{table}{0}

\section{Experiment details}\label{append:exp_details}
We conducted our experiments using four 80GB NVIDIA A100 GPUs. 

In our learning strategy for expert merging, we used SGD as the optimizer to learn the coefficients for expert merging, with the learning rate set to $1 \times 10^{-3}$. We randomly sample $128$ samples from the calibration set (C4) and partition them into a $3:1$ ratio as the training and evaluation sets, respectively.  The coefficients were initialized as identity matrices and optimized over $50$ epochs until they converged. 

For baseline methods and models, all usage and distribution comply with the terms of their license, \textit{i.e.}, Mixtral (Apache License 2.0), Deepseek (MIT license), Qwen (Tongyi Qianwen license), lm evaluation harness (MIT license). We use Copilot to help with debugging and coding.

\section{Evaluation on the expert similarity}
\label{sec:apped_results}

\begin{table*}[]
\centering
\caption{Results on pruning the Mixtral-8x7B and Mixtral-8x22B from 8 experts to 6 and 4 experts in each MoE layer. We present the results of our four strategies, namely 1) Vectorized and surrogate $\theta$: prunable experts discovery using vectorized  or surrogate weights and merging with uniform coefficients;  2) Learn: prunable experts discovery using vanilla data and learn coefficients to merge based on \cref{eq:loss_learn_alpha}; 3) Max: maintaining the expert in each discovered group with the maximum visiting frequency.  The first and second columns respectively indicate the results on pruned model with $6$ and  $4$ experts.}
\label{tab:main_result_7b}
\resizebox{\textwidth}{!}{
\begin{tabular}{cc|cccccccc}
\toprule
\multicolumn{2}{c|}{Dataset} & \multicolumn{4}{c}{MMLU}                   & \multirow{2}{*}{BoolQ} & \multirow{2}{*}{OpenBookQA} & \multirow{2}{*}{RTE} & \multirow{2}{*}{Average} \\
\multicolumn{2}{c|}{Method}  & humanities & social science & stem & Other &                        &                             &                      &                          \\\hline
\multicolumn{2}{c|}{\gray{Mixtral-8x7B}}  &            \gray{60.5}         & \gray{77.8}               & \gray{58.9}     &    \gray{74.2}   &      \gray{85.4}                  &         \gray{34.4}                   & \gray{71.1}                     &    \gray{66.0}                
 \\\hline
\multirow{2}{*}{\textbf{Model-centric}}    & \textbf{Vectorized}  & 54.4/\textbf{48.1}       &    70.2/\textbf{58.5}            &  51.8/\textbf{45.2}    &   66.8/\textbf{55.2}    &                85.6/83.7        &     31.4/26.2        &       68.9/62.4            &  61.3/54.2                        \\& \textbf{Surrogate}  & 56.8/47.1       &    69.7/56.4            &  50.9/42.2    &   66.0/55.0    &                \textbf{86.9}/\textbf{83.8}        &     32.6/27.0        &       68.5/64.6            &  61.6/53.7                        \\\hdashline
\multirow{3}{*}{\textbf{Data-centric}}
                           & \textbf{Learn} & 56.0/48.0                          &     \textbf{73.1}/57.0 &    \textbf{52.4}/43.3   &          \textbf{68.2}/54.6              &    {86.4}/83.3           &         31.4/28.5      &         \textbf{69.3}/\textbf{67.1}             &     62.4/\textbf{54.5}\\
                         &
                         \textbf{Max}  & 56.4/47.6       &    71.9/58.6            &  52.0/42.9    &   66.9/55.7    &                85.1/82.8        &     \textbf{35.2}/\textbf{28.8}        &       70.4/66.8            &  \textbf{62.6}/53.7\\
                         &
                         \textbf{Uniform}  & 56.2/47.8       &    72.7/57.2            &  52.1/42.7    &   68.1/54.0    &                85.6/{83.2}        &     32.8/28.6        &       68.6/66.5            &  62.3/54.3\\\hline\hline

\multicolumn{2}{c|}{\gray{Mixtral-8x22B}}  &            \gray{68.6}         & \gray{84.1}               & \gray{67.1}     &    \gray{78.7}   &      \gray{87.9}                  &         \gray{0.358}                   & \gray{71.2}                     &    \gray{70.4}                   \\\hline
   \multirow{2}{*}{\textbf{Model-centric}}                             & \textbf{Vectorized} &26.7/24.2 & 28.5/21.7 & 26.9/21.3 & 31.7/23.8 & 62.0/53.6 & 19.6/11.4 & 52.0/53.1 & 35.3/29.9\\
   & \textbf{Surrogate}  &\textbf{63.7}/\textbf{58.1} & \textbf{80.0}/\textbf{72.5} & \textbf{62.1}/\textbf{54.3} & \textbf{75.6}/\textbf{68.3} & 88.0/\textbf{85.2} & 34.6/31.2 & 69.0/\textbf{68.6} & \textbf{67.6}/\textbf{62.6} \\\hdashline
   \multirow{3}{*}{\textbf{Data-centric}}
   & \textbf{Learn} & 61.4/57.9 & 78.3/70.1 & 61.2/51.4 & 72.8/65.0 & \textbf{88.2}/84.9 & 35.6/32.7 & 70.5/67.3    & 66.9/61.3\\
   &\textbf{Max} & 56.8/47.1       &    69.7/56.4            &  50.9/42.2    &   66.0/55.0    &                86.9/{83.8}        &     32.6/27.0        &       68.5/64.6            &  61.6/53.7 
   \\
   &
                         \textbf{Uniform}  & 62.3/57.8       &    78.5/69.7            &  60.2/51.3    &   73.4/64.2    &                87.6/{83.1}        &     \textbf{35.8}/\textbf{33.2}        &       71.1/68.1            &  67.0/61.1\\\bottomrule
\end{tabular}}
\end{table*}

\begin{figure*}
    \centering
    \includegraphics[width=\linewidth]{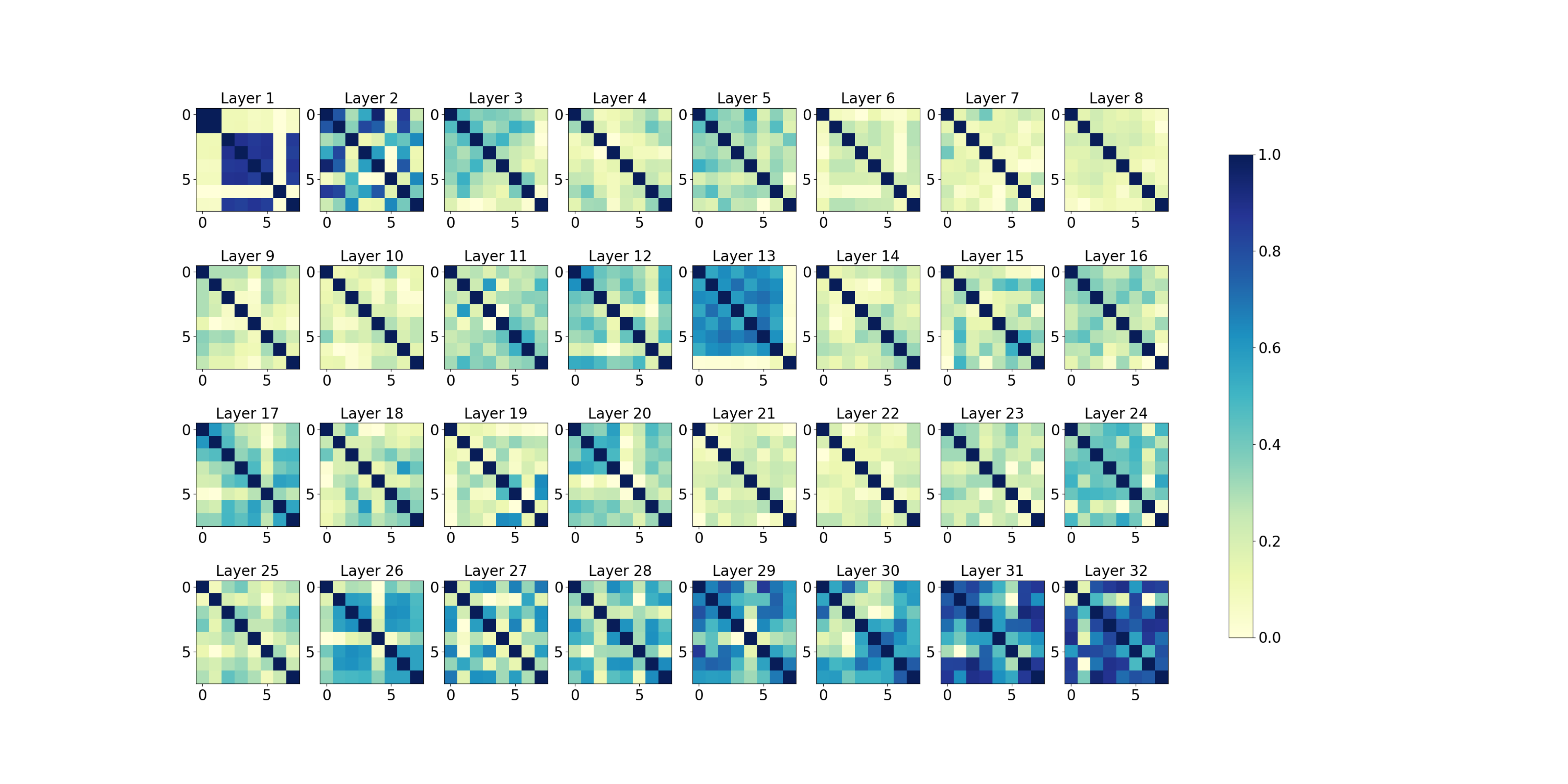}
    \caption{Evaluation of the expert similarity for different MoE layers in Mixtral-8x7B under the linear kernel-based CKA criteria.}
    \label{fig:full_cka_7b }
\end{figure*}

\begin{figure*}
    \centering
    \includegraphics[width=\linewidth]{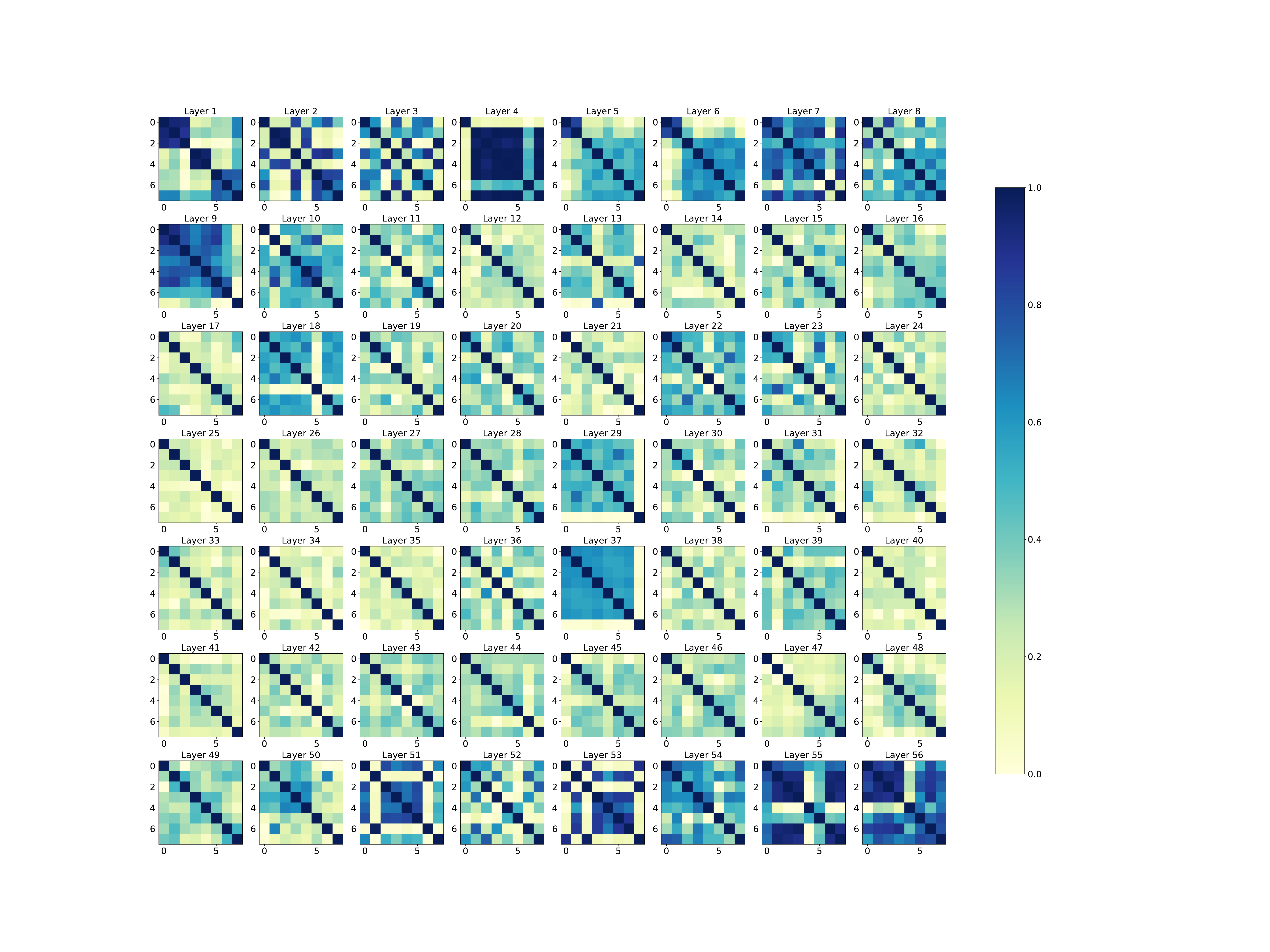}
    \caption{Evaluation of the expert similarity for different MoE layers in Mixtral-8x22B under the linear kernel-based CKA criteria.}
    \label{fig:full_cka_22b}
\end{figure*}

Following the same setting in \cref{sec:motivation}, we conduct experiments on evaluating the expert similarity in all MoE layers of Mixtral-8x7B and Mixtral-8x22B. The results are respectively depicted in \cref{fig:full_cka_7b } and \cref{fig:full_cka_22b}.
We summarize the observation as follows,
\begin{itemize}
    \item Most MoE layers in the two Mixtral models contain significant expert redundancy.
    \item The most redundant MoE layers are located in the first and last several layers, while the experts in the intermediate MoE layers learn more diverse features.
\end{itemize}

\section{Enumeration on the expert pruning}
\label{append_sup:enum}
\begin{figure*}
    \centering
    \includegraphics[width=0.6\linewidth]{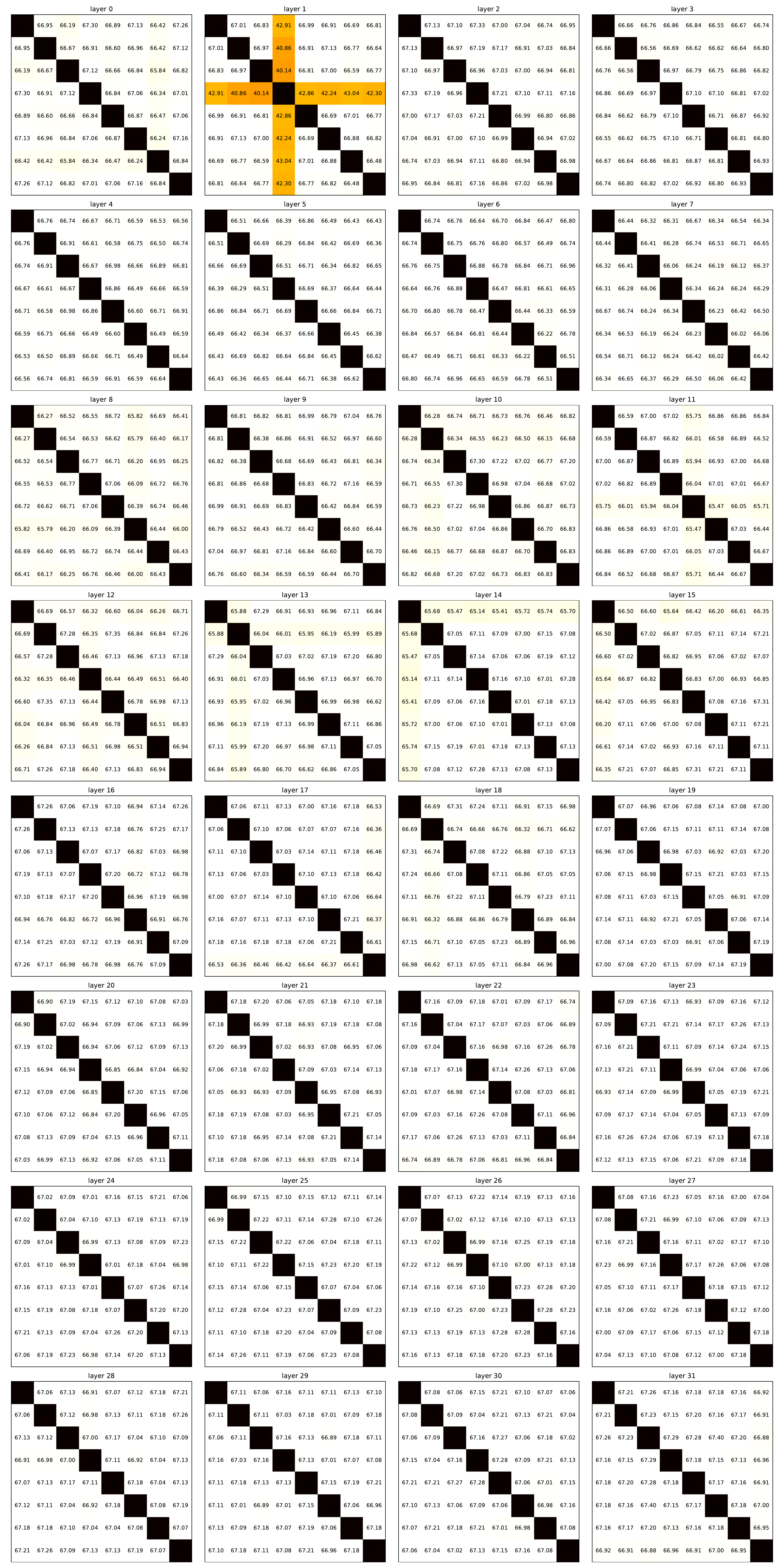}
    \caption{Enumeration on dropping two experts.}
    \label{fig:enum_6}
\end{figure*}

We present the full greedy search result on Mixtral-8x7B in \cref{fig:enum_6}. In detail, we first enumerate all the possible combinations of dropping $2$ experts layer by layer, and then evaluate the model on the MMLU task. For example, $66.19$ in the first row and third column in layer 0 indicates performance while dropping the first and third expert in layer 0 of the Mixtral-8x7B model.

Although dropping most of the combinations on different layers only leads to a minor performance drop, we can notice that it could cause the model to crash when the fourth expert in layer 1 is involved during the pruning process.

\section{Results on different strategies for pruning}
\label{append_sup:more_results}  

While we report the best performance using our data-centric and model-centric strategies to prune the MoE models, we detail more results on pruning the Mixtral-8x7B and Mixtral-8x22B. 
\begin{enumerate}
    \item For model-centric stratgies, we show the results with \textbf{vectorized} weight and \textbf{surrogate} weight strategies to discover the similar experts. We uniformly merge the experts and their routers for model pruning. In other words, \textit{the model-centric strategies differ at the discovery stage}.
    \item For data-centric strategy, after we discover similar experts, we respectively use the learning strategy to merge the experts with weights (\textbf{Learn}), only maintain the expert with the maximum visiting frequency (\textbf{Max}), or uniformly merging different experts (\textbf{Uniform}). Thus, in this experiment, \textit{the data-centric strategies differ at the merging stage}. 
\end{enumerate}

From the results of our strategies, we can observe the following:
\begin{enumerate}
    \item \textit{Both the Vectorized  and Surrogate strategies surpass all other data-involved pruning baseline methods.} This suggests that \textit{weights already encode valuable data information, which can be utilized to group experts for pruning}.
    \item \textit{When pruning relatively small models (Mixtral-8x7B), the inclusion of data in the pruning process improves candidate selection for merging and pruning, leading to better performance compared to using only model weights.} While the learning strategy offers a slight improvement in average performance, it comes at a higher computational cost compared to using uniform coefficients for merging.
    \item \textit{For larger models (Mixtral-8x22B), the model-centric method outperforms the data-centric method.} We argue that this is due to overfitting to the small calibration dataset during pruning, especially when dealing with a large number of parameters. The small calibration dataset cannot approximate the distribution of the pre-training dataset effectively. This is evident from the significant performance drop observed when using the Max strategy for pruning Mixtral-8x22B.
\end{enumerate}

\section{Full results on pruning Deepseek and Qwen}
\begin{figure*}
    \centering
    \includegraphics[width=\linewidth]{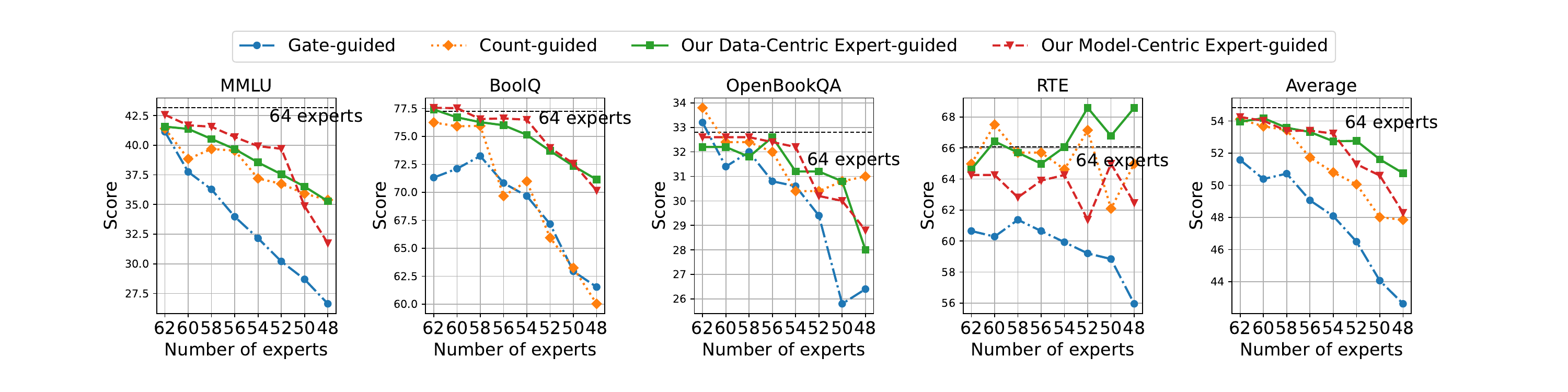}
    \caption{Results on pruning Deepseek-MoE-16B.}
    \label{fig:deepseek_moe_full}
\end{figure*}

\begin{figure*}
    \centering
    \includegraphics[width=\linewidth]{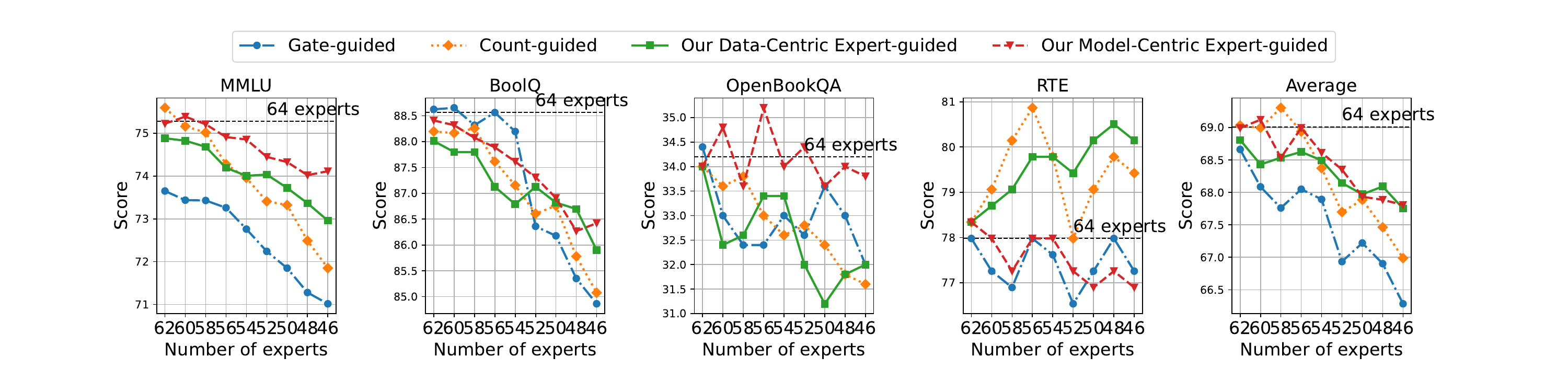}
    \caption{Results on pruning Qwen2-57B-14A.}
    \label{fig:qwen_moe_full}
\end{figure*}

We present the full results of pruning Deepseek-MoE-16 and Qwen2-57B-14A in \cref{fig:deepseek_moe_full} and \cref{fig:qwen_moe_full}. It is evident that our proposed data-centric and model-centric strategies outperform all baseline methods in most test cases. Additionally, when examining different evaluation tasks, the results on MMLU show a more reliable and consistent trend as the number of pruned experts increases. However, for smaller tasks such as the RTE dataset, we observe some randomness in the evaluation results due to the limited dataset size.We did not include the enumeration-based method in our comparison, as it is time-consuming and difficult to complete within a limited timeframe, especially when the number of experts is large in these models.

\section{Empirical Analysis on Expert Hints}\label{append_hint}
We analyze the change of expert hints on the calibration dataset C4, where the expert hint refers to the visiting frequency of the expert on the calibration set.

\begin{figure}[htbp]
    \centering
    \includegraphics[width=\linewidth]{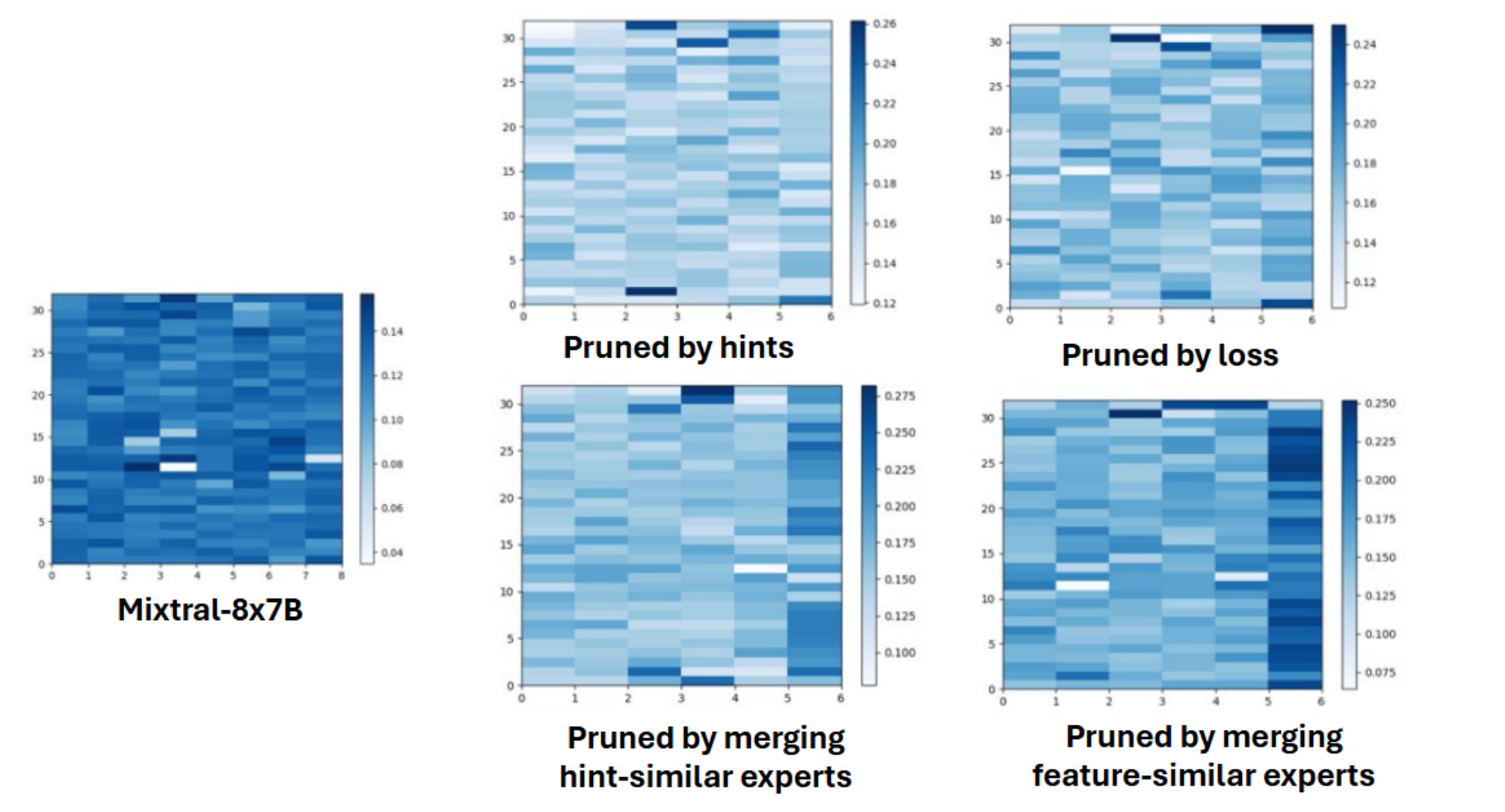}
    \caption{Statistics of the visiting frequency for all experts in different MoE layers. }
    \label{fig:frequency_cmp}
\end{figure}

The statistic results are shown in \cref{fig:frequency_cmp}.   We can see that in most MoE layers in Mixtral-8x7B, many pairs of experts have similar hints. In contrast, pruning differentiates the hints of experts. Compared to directly using hints as the pruning goal, using expert knowledge as the pruning criterion results in more significant changes in hints. Additionally, the merging operation on the gate aggregates the new expert (last column) more tokens, further increasing the hint differences.

\section{Pseudo-code implementation}\label{append_code}
\begin{algorithm}
	\caption{Pruning algorithm for given MoE layer}
	\begin{algorithmic}[1]
		\Require  The set of experts $\mathcal{F}=\{f_{1}, f_{2},...,f_{N}\}$ in the given MoE layer to prune, each expert $f_{i}$ consisting of three layers parameterized by $\theta^i_1, \theta^i_2, \theta^i_3$, respectively, the number of experts to reserve $r$, the calibration dataset $\mathcal{D}$, strategy s.
		\Ensure Pruned MoE layer with reduced experts $\{f_{1},f_{2},...,f_{r}\}$.
		\Function{MoE pruning}{MoE layer $\mathcal{F}$, number of experts $r$ to reserve}
		\State Initialize $\mathcal{G}=\{\mathcal{V},\mathcal{E}, \mathcal{A}\}$.
            \For {each $i$ in $N$}
            \If{s==Data-centric}
                 \State $\mathcal{R}(f_{i}) = f_{i}(\text{Mixup}(\mathcal{D}))$.
            \ElsIf{s==Vectorized weights}
                 \State $\mathcal{R}(f_i) = \{\theta^i_1, \theta^i_2, \theta^i_3\}$ 
            \ElsIf{s==Surrogate weights}
                 \State $\mathcal{R}(f_i) = \theta^i_2(\theta^i_1\cdot\theta^i_3$) 
            \EndIf
            \EndFor
            \For {each $i$ in $N$}  
            \For {each $j$ in $N$}
		\State $\mathcal{A}(\mathcal{E}_{ij})=\text{CKA}(\mathcal{R}(f_{i}),\mathcal{R}(f_{j}))$
		\EndFor
             \EndFor
            \State Optimize \cref{eq:find} on $\mathcal{G}$  to find $r$ subgroup of experts.
            \State Merge experts with their routers clustered within the same subgroup based on \cref{eq:model_soup}. 
		\State \Return{Pruned $\mathcal{F}$} 
		\EndFunction
	\end{algorithmic}
\end{algorithm}

Solving \cref{eq:find} is an NP-hard problem. To obtain an approximate solution, two methods can be employed. One approach is to use spectral clustering~\citep{ng2001spectral}, while the other is a greedy strategy that iteratively merges the pair of experts with the highest similarity to construct the pruned expert graph. Although both methods yield comparable performance, the greedy approach is significantly more efficient, particularly when dealing with large language models that have many experts per MoE layer. In practice, for data-centric method,  we also observe that using cosine similarity performs on par with centered kernel alignment (CKA), while being more computationally efficient for measuring expert similarity. At the same time, the mean-squared error also works well, which we adopt in the model-centric method.

\section{Discussion on data- and model-centric methods}\label{append:discuss_two_strategies}
The data-centric method is a natural extension of the paper’s starting point. However, its implementation requires access to a small portion of the pre-training dataset or a surrogate dataset with a similar distribution, which may pose practical challenges.

On the one hand, achieving good distributional similarity between the surrogate and real datasets is indeed crucial for the success of the data-centric method. Without such similarity, the method can fail entirely. In our trials, we tried using random inputs to identify similar experts and prune the MoE model, which led to the crashed performance of the pruned experts.

On the other hand, while surrogate datasets are generally simpler than real-world datasets, they still capture critical characteristics learned by different experts. Notably, no pre-trained data is publicly available for the models we study, including Mixtral, DeepSeek, and Qwen. However, leveraging a small portion of the open-source pre-trained dataset C4 still yields comparable performance on diverse benchmarks, as shown in our results.

In contrast, the model-centric method identifies similar experts directly by analyzing the model's weights, bypassing the need for any data. This approach builds on the observation that training on the same dataset produces similar weight distributions, which can serve as a surrogate for predicting outputs given the same inputs. In the context of MoEs, when the gating mechanism consistently routes similar tokens to a group of experts, those experts tend to exhibit similar weights. Therefore, the model-centric strategy aligns with the paper's initial data-dependent premise by offering a complementary perspective on expert similarity without relying on external datasets.


\end{document}